\documentclass[conference]{IEEEtran}
\IEEEoverridecommandlockouts
\usepackage{cite}
\usepackage{amsmath,amssymb,amsfonts}
\usepackage{algorithmic}
\usepackage{graphicx}
\usepackage{textcomp}
\usepackage{xcolor}

\usepackage{caption}
\def\BibTeX{{\rm B\kern-.05em{\sc i\kern-.025em b}\kern-.08em
    T\kern-.1667em\lower.7ex\hbox{E}\kern-.125emX}}
\begin{document}

\title{
Adaptive Routing for Efficient Diffusion Transformer-Based PNI Prediction}

\author{
\IEEEauthorblockN{
Youngung Han$^{1,3}$, Dohyun Kweon$^{2,3}$, Kyeonghun Kim$^{3}$, Hyunsu Go$^{1}$, Jina Jeong$^{1}$, Suah Park$^{1}$, \\
Induk Um$^{4}$, Junga Kim$^{1}$, Anna Jung$^{1}$, Yului Jeong$^{1}$, Sungha Park$^{1,5}$, Jinyong Jun$^{1}$\\
Pa Hong$^{6}$, Woo Kyoung Jeong$^{7}$, Won Jae Lee$^{6}$, Ken Ying-Kai Liao$^{8}$, Hyuk-Jae Lee$^{1}$, Nam-Joon Kim$^{1,\dag}$
}

\vspace{0.6em}

\IEEEauthorblockA{
$^{1}$Seoul National University, Seoul, Republic of Korea \quad
$^{2}$Kyung Hee University, Seoul, Republic of Korea
}

\IEEEauthorblockA{
$^{3}$OUTTA, Seoul, Republic of Korea \quad
$^{4}$Chung-Ang University, Seoul, Republic of Korea
}

\IEEEauthorblockA{
$^{5}$Seoul National University School of Medicine, Seoul, Republic of Korea
}

\IEEEauthorblockA{
$^{6}$Samsung Changwon Hospital, Changwon, Republic of Korea \quad
$^{7}$Samsung Medical Center, Seoul, Republic of Korea
}

\IEEEauthorblockA{
$^{8}$NVIDIA AI Technology Center, Taipei, Taiwan
}

\IEEEauthorblockA{
$^{\dag}$Corresponding author: \texttt{knj01@snu.ac.kr}
}
}

\maketitle

\begin{abstract}

Perineural invasion (PNI) is a critical prognostic factor in cholangiocarcinoma. However, its preoperative prediction from magnetic resonance imaging (MRI) remains challenging due to subtle imaging features that extend beyond tumor boundaries into surrounding regions.
Conventional convolutional neural networks are limited in capturing long-range spatial dependencies. Transformer-based architectures improve global modeling of volumetric MRI by aggregating spatially distributed contextual cues, yet capturing subtle and noise-sensitive patterns in peritumoral regions remains challenging. Diffusion-based classifiers offer an alternative formulation by leveraging denoising-based class scoring to better capture such subtle patterns. However, these approaches introduce substantial computational overhead due to the combination of transformer-based modeling and iterative denoising processes. To address these challenges, we formulate PNI prediction as a diffusion-based classification problem and implement the denoising network using a transformer-based representation. To improve computational efficiency, we introduce adaptive routing across attention heads, spatial tokens, and MLP width. Experimental results demonstrate that the proposed approach achieves an AUC of 0.731 with 257.57 GFLOPs.
\end{abstract}


\section{Introduction}
Perineural invasion (PNI) is a critical prognostic factor closely associated with tumor aggressiveness and poor clinical outcomes~\cite{wei2022prognostic}. 
Preoperative prediction of PNI from MRI may assist in surgical planning and inform neoadjuvant treatment strategies; however, it remains challenging due to subtle and spatially diffuse imaging features that often extend beyond tumor boundaries~\cite{han2026losa,han2026mma,han2026neonet}. PNI involves tumor spread along and around nerve structures, resulting in weak and spatially distributed patterns that are difficult to detect~\cite{liu2024noninvasive, liebig2009perineural,conti2026perineural,li2020perineural,wei2022prognostic}.
CNN-based models primarily rely on local receptive fields~\cite{simonyan2014very}, which may limit their ability to capture long-range spatial dependencies across slices and surrounding peritumoral regions. Subtle peritumoral cues are often weak and spatially distributed beyond localized regions, making them difficult to capture with purely local representations.

Transformer-based architectures model global interactions via self-attention~\cite{dosovitskiy2020image,hatamizadeh2022unetr,chen2021transunet}, aggregating dispersed contextual cues. However, subtle peritumoral cues remain sensitive to noise and ambiguity, especially in medical imaging with low signal-to-noise ratios and inter-patient variability~\cite{isensee2021nnu}. Diffusion-based classifiers have therefore been explored to improve robustness to subtle and noisy patterns~\cite{chen2023robust, clark2023text}.

Diffusion models probabilistically model data distributions~\cite{song2021ddim,ho2020denoising} and can be adapted for classification by comparing class-conditional denoising or reconstruction errors~\cite{clark2023text}. To combine denoising-based classification with long-range volumetric MRI modeling, we adopt a transformer-based denoising backbone~\cite{peebles2023scalable,han2025foscu, kim20263d}.


Despite these advantages, applying diffusion transformers to 3D MRI introduces substantial computational overhead due to the large number of tokens and repeated denoising steps, posing practical challenges for hardware-efficient deployment in clinical settings. 
While dynamic token selection has been explored for efficient vision transformers~\cite{rao2021dynamicvit}, identifying relevant tokens in volumetric medical imaging requires awareness of local anatomical context around tumor boundaries. 

To address this, we propose Diffusion Transformer with Routing for Classification (DiT-RC), 
an efficient diffusion transformer classifier with adaptive routing across attention heads, 
spatial tokens, and MLP width for volumetric MRI.
The proposed design dynamically allocates computation based on input complexity, enabling more effective identification of clinically relevant regions around tumor boundaries.

Our contributions are summarized as follows:
\begin{itemize}
    \item We propose DiT-RC, a diffusion transformer-based framework for PNI prediction from 3D MRI, leveraging denoising-based classification.
    \item We introduce dynamic routing across attention heads, spatial tokens, and MLP width to enable adaptive computation.
    \item We incorporate a lightweight convolutional module for improved token importance estimation around tumor boundaries.
\end{itemize}




\section{Proposed Method}
\begin{figure*}[t]
\centering
\includegraphics[width=1\textwidth]{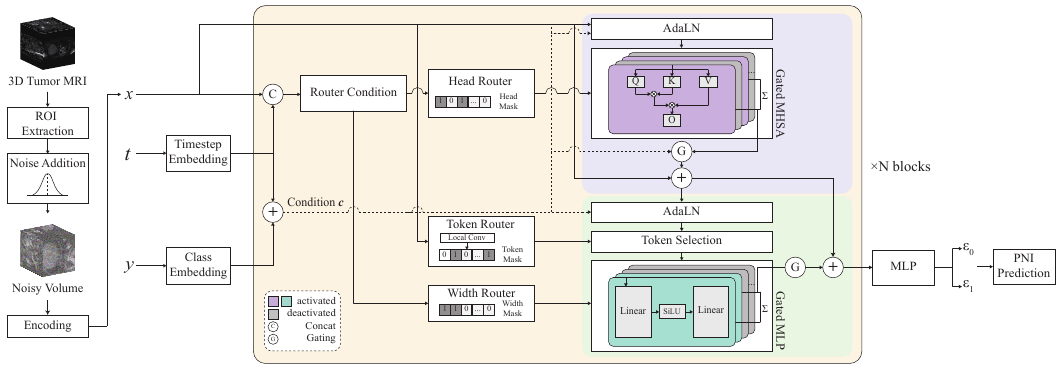}
\caption{Overview of DiT-RC. 
The architecture consists of a stack of N transformer blocks (N = 6 in our implementation).}
\label{fig:framework}
\end{figure*}


\subsection{Diffusion Transformer for Classification}

We employ a diffusion-based transformer~\cite{li2022dit} classifier to predict
perineural invasion (PNI) from tumor-centered 3D MRI volumes.
Given a clean input volume $x_0$, a forward diffusion process generates a noisy sample $x_t$ at timestep $t$:
\begin{equation}
x_t = \sqrt{\bar{\alpha}_t}\, x_0 + \sqrt{1 - \bar{\alpha}_t}\, \epsilon
\end{equation}
where $\epsilon \sim \mathcal{N}(0, I)$ and $\bar{\alpha}_t$ denotes the cumulative noise schedule.

The noisy volume $x_t$ is partitioned into non-overlapping 3D patches and embedded into a token sequence $X_t$, which is processed by a transformer backbone conditioned on both the timestep and a class hypothesis. Specifically, the model estimates the noise 

\begin{equation}
\hat{\epsilon}_\theta(x_t,t,y)
\end{equation}

where $y \in \{0,1\}$ denotes the class hypothesis for PNI status.
During inference, the same noisy input is evaluated under each class condition, and the prediction is determined by comparing the corresponding noise reconstruction errors $m_y = \|\hat{\epsilon}_\theta(x_t, t, y) - \epsilon\|_2^2$, with $\hat{y} = \arg\min_y m_y$.
Compared with a conventional discriminative classifier, this formulation
leverages noise-perturbed inputs across multiple timesteps and encourages
the model to learn representations that remain informative under stochastic
corruption.

\subsection{Adaptive Computation}
To improve efficiency, we introduce adaptive routing within each transformer block, 
controlling attention-head activation, spatial token selection, and MLP width for 
the token sequence $X\in\mathbb{R}^{N\times C}$. The routing condition is defined as 
$r=[t_{\mathrm{emb}};x_{\mathrm{global}}]$, where 
$x_{\mathrm{global}}=\frac{1}{N}\sum_{i=1}^{N}X_i$. 
It is used for attention-head and MLP-width routing, while token routing uses 
token features with local 3D context. AdaLN~\cite{guo2022adaln,perez2018film} modulates each block with $c=t_{\mathrm{emb}}+y_{\mathrm{emb}}$.


\subsubsection{Attention Head Routing}
Multi-head self-attention models diverse feature interactions through multiple attention heads. 
Given routing condition r, a lightweight router predicts head-wise scores $s^{\mathrm{head}}\in\mathbb{R}^{H}$ from the routing condition $r$. 
These scores are converted into a binary routing mask $m^{\mathrm{head}}\in\{0,1\}^{H}$ using a Gumbel--Sigmoid estimator~\cite{jang2016categorical}.

The routed attention output is defined as
\begin{equation}
\mathrm{MHSA}(X) = \sum_{h=1}^{H} m_h^{\mathrm{head}} \cdot \mathrm{Attn}_h(X)
\label{eq:mhsa}
\end{equation}

where $H$ denotes the number of attention heads and $\mathrm{Attn}_h(\cdot)$ is the attention computation over head $h$.

\subsubsection{Spatial Token Routing}
Volumetric MRI inputs yield a large number of spatial tokens, many of which contain limited information for PNI prediction; we therefore introduce spatial token routing to focus computation on informative regions using local spatial context.
Specifically, token scores $s^{\mathrm{tok}} \in \mathbb{R}^N$ are computed by combining each token's own features with local spatial context extracted via a lightweight 3D convolution (kernel size $3\!\times\!3\!\times\!3$) applied over the spatial token grid, which are then converted into a binary mask $m^{\mathrm{tok}} \in \{0,1\}^N$.

The routed token representation is
\begin{equation}
X' = m^{\mathrm{tok}} \odot X 
\end{equation}
where $\odot$ denotes token-wise masking. For notational simplicity, token routing is written as masking, while the actual implementation performs sparse token selection before the MLP block so that only selected tokens participate in the feed-forward computation.

\subsubsection{MLP Width Routing}
The transformer MLP expands token features into a high-dimensional space, making it a major source of computational cost. Given the routing condition $r$, a lightweight router predicts width scores $s^{\mathrm{mlp}}\in\mathbb{R}^{G}$, which are converted into a binary width mask $m^{\mathrm{mlp}}\in\{0,1\}^{G}$.

Let $Z=\phi(W_1X')$ denote the hidden activation of the MLP, where $W_1$ is the first projection matrix and $\phi(\cdot)$ is the non-linear activation function. 
Width routing applies channel-group masking to the hidden representation,
\begin{equation}
Z' = m^{\mathrm{mlp}} \odot Z 
\end{equation}
and the final MLP output is computed from $Z'$. 
Since the MLP is evaluated only for tokens selected by spatial token routing, token routing and width routing jointly reduce the feed-forward computation.

These routing mechanisms focus computation on clinically relevant regions while reducing overall cost by dynamically adapting the computation pathway to the noisy input and diffusion timestep.
\subsection{Training Objective}
The proposed model is trained with a joint objective that combines
diffusion-based classification with a routing budget constraint:
\begin{equation}
    \mathcal{L}_{\text{total}} = \mathcal{L}_{\text{cls}} + \lambda_{\text{budget}}\mathcal{L}_{\text{budget}}
    \label{eq:total}
\end{equation}
where $\mathcal{L}_{\text{cls}}$ denotes the diffusion classification loss and
$\mathcal{L}_{\text{budget}}$ regulates the adaptive routing behavior. The budget loss is only applied when routing is enabled, as it requires controllable computation through routing decisions. 
 
For a noisy input $x_t$, we compute reconstruction errors $m_y$ for $y \in \{0,1\}$ as defined in Sec. $\text{II}$-B and convert them into classification logits $\mathbf{l} = \tau \cdot [-m_0, -m_1]$, where $\tau$ is a learnable temperature parameter.

To control computational cost, the effective computation
ratio $r_{\text{eff}}$ is defined as:
\begin{equation}
    r_{\text{eff}} = w_{\text{attn}}k_{\text{attn}} + w_{\text{mlp}}(k_{\text{tok}}k_{\text{mlp}})
    \label{eq:reff}
\end{equation}

where $r_{\text{eff}}$ approximates the relative computational cost in terms of FLOPs, providing a hardware-agnostic proxy for efficiency~\cite{cai2019once, rao2021dynamicvit}, and $k_{\text{attn}}, k_{\text{tok}}, k_{\text{mlp}}$ denote the average keep ratios of attention heads, spatial tokens, and MLP hidden channels, respectively.
 
 The classification and budget losses are then given by:
\begin{equation}
    \mathcal{L}_{\text{cls}} = \mathcal{L}_{\text{mse}}^{(y)} + \alpha \mathcal{L}_{\text{CE}}, \quad
    \mathcal{L}_{\text{budget}} = (r_{\text{eff}} - \lambda_{\text{target}})^2
    \label{eq:losses}
\end{equation}
 
The coefficient $\lambda_{\text{target}}$ is gradually increased during early
training to stabilize optimization. This objective jointly learns diffusion
classification and dynamic routing policies, enabling the model to balance
predictive performance and computational efficiency.

\section{Experiment Results}
\subsection{Dataset and Implementation Details}

The dataset consists of 155 patients (61 PNI-positive and 94 PNI-negative) collected over a 10-year period at Samsung Medical Center. T2-weighted MRI scans were used for all experiments. Each patient volume was cropped to a tumor-centered region of size $96\!\times\!96\!\times\!48$ and divided into non-overlapping 3D patches of size $6\!\times\!6\!\times\!6$, forming a $16\!\times\!16\!\times\!8$ token grid. Tumor masks were manually delineated using 3D Slicer and used only for 
tumor-centered ROI extraction, not as model inputs.

We performed 5-fold stratified cross-validation at the patient level to avoid data leakage across scans from the same patient. The model was trained using AdamW with a learning rate of $8\times10^{-5}$, weight decay of $0.05$, and a batch size of $4$. All experiments were conducted on a single NVIDIA RTX 3090 GPU using mixed-precision training.

During training, diffusion timesteps were randomly sampled from a 1000-step schedule. During inference, predictions were obtained by averaging logits over multiple timesteps ($t\in\{200,350,500,650,800\}$) and random seeds, following standard diffusion-based classification protocols to reduce stochastic variance. The 95\% confidence interval was estimated by bootstrapping patient-level predictions. FLOPs and latency were reported per patient prediction under the five-timestep and two-class protocol, excluding repeated noise ensembling. For routed models, FLOPs were computed using the activated heads, selected tokens, and MLP channel groups. For robustness evaluation, additive Gaussian noise with standard deviations $\sigma \in \{0.1, 0.3, 0.5\}$ was applied after per-volume min–max normalization, using pre-generated corrupted test sets shared across all models.

\subsection{Quantitative and Qualitative Results}

\begin{table}[t]
\centering
\caption{Classification performance (AUC) under varying noise levels for PNI prediction.}
\label{tab:performance}
\begin{tabular}{l c c c c}
\hline
Model & Clean & NL & NM & NH \\
\hline
ResNet-18~\cite{he2016deep} & 0.675 & 0.630 & 0.573 & 0.475 \\
DenseNet-121~\cite{huang2017densely} & 0.687 & 0.643 & 0.573 & 0.527 \\
EfficientNet~\cite{tan2019efficientnet} & 0.680 & 0.645 & 0.612 & 0.500 \\
Vision Transformer~\cite{dosovitskiy2020image} & 0.700 & 0.654 & 0.633 & 0.610 \\
Swin Transformer~\cite{liu2021swin} & 0.710 & 0.700 & 0.649 & 0.615 \\
Diffusion Classifier~\cite{ho2020denoising} & 0.692 & 0.692 & 0.680 & 0.672 \\
\hline
\textbf{DiT-RC} & \textbf{0.731} & \textbf{0.730} & \textbf{0.723} & \textbf{0.710} \\
\hline
\end{tabular}
\end{table}

Table~\ref{tab:performance} summarizes the AUC performance under clean and noisy inputs. 
CNN-based models show relatively lower performance, while transformer-based 
models improve clean AUC but degrade under stronger noise. Diffusion-based 
models are more robust to noise, and DiT-RC achieves the best AUC across all 
settings. DiT-RC achieved an AUC of $0.731$ on clean inputs 
(bootstrap 95\% CI: $0.66$--$0.80$) and showed five-fold standard deviations of 
$0.036/0.038/0.041/0.046$ for Clean/NL/NM/NH.

Table~\ref{tab:efficiency_unet_ours} compares our method with the same diffusion classifier implemented using a U-Net backbone based on MONAI and DDPM~\cite{ho2020denoising}, which represents a standard CNN-based diffusion approach. While the U-Net-based diffusion classifier achieves lower computational cost in terms of FLOPs and latency, our method attains a higher AUC under the same per-prediction evaluation setting. This suggests that transformer-based denoising better captures PNI-related volumetric context, while introducing additional computational cost compared with the CNN-based diffusion baseline. 

\begin{table}[t]
\centering
\small
\setlength{\tabcolsep}{6pt}
\caption{Comparison with Diffusion Classifier.}
\label{tab:efficiency_unet_ours}
\begin{tabular}{lccc}
\hline
Model & FLOPs (G) & Lat. (ms) & AUC \\
\hline
Diffusion Classifier~\cite{ho2020denoising} & 151.75 & 127.10 & 0.692 \\
\textbf{DiT-RC} & \textbf{257.57} & \textbf{140.47} & \textbf{0.731} \\
\hline
\end{tabular}
\end{table}

\begin{figure}[t]
\centering
\includegraphics[width=0.9\linewidth]{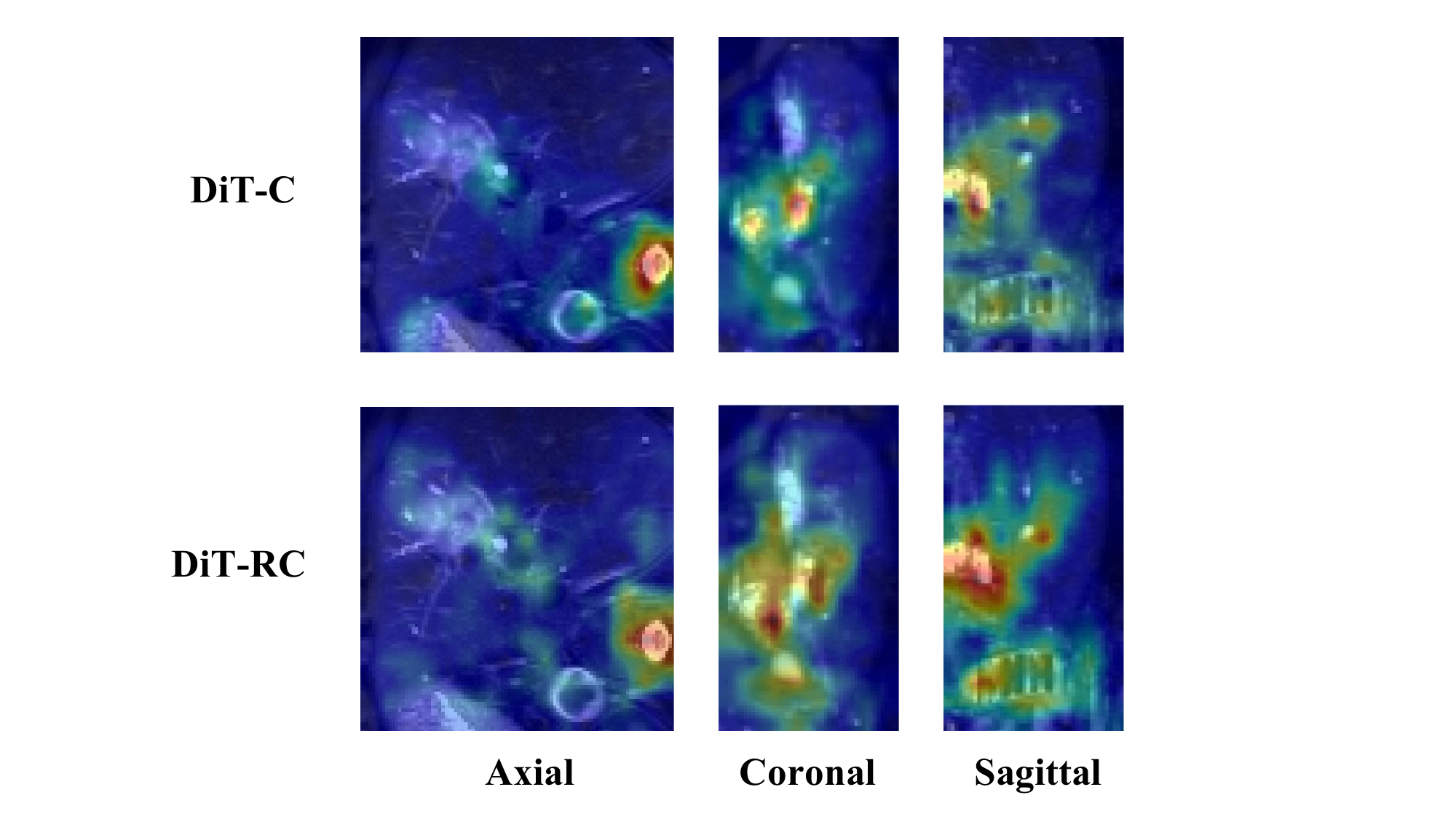}
\caption{Grad-CAM visualization for PNI prediction.}
\label{fig:gradcam}
\end{figure}

Figure~\ref{fig:gradcam} shows Grad-CAM~\cite{selvaraju2017grad} visualizations across axial, coronal, and sagittal views. 
The red regions indicate areas associated with PNI-positive predictions. The proposed model produces activation patterns similar to those of its non-routing variant (DiT-C), while reducing computational cost.
\subsection{Efficiency Analysis}

\begin{table}[t]
\centering
\small
\setlength{\tabcolsep}{6pt}
\caption{Efficiency comparison between DiT-RC and its non-routing variant (DiT-C).}
\label{tab:efficiency}
\begin{tabular}{lccc}
\hline
Model & FLOPs (G) & Lat. (ms) & AUC \\
\hline
DiT-C & 418.79 & 319.32 & 0.733 \\
\textbf{DiT-RC} & \textbf{257.57} & \textbf{140.47} & \textbf{0.731} \\
\hline
\end{tabular}
\end{table}

Transformer-based diffusion classifiers are computationally expensive compared to CNN-based approaches. To address this, we introduce a routing mechanism that dynamically reduces redundant computation. Table~\ref{tab:efficiency} shows that routing significantly reduces FLOPs from 418.79G to 257.57G and latency from 319.32 ms to 140.47 ms, while maintaining comparable performance.

\begin{figure}[!t]
\centering
\includegraphics[width=\columnwidth]{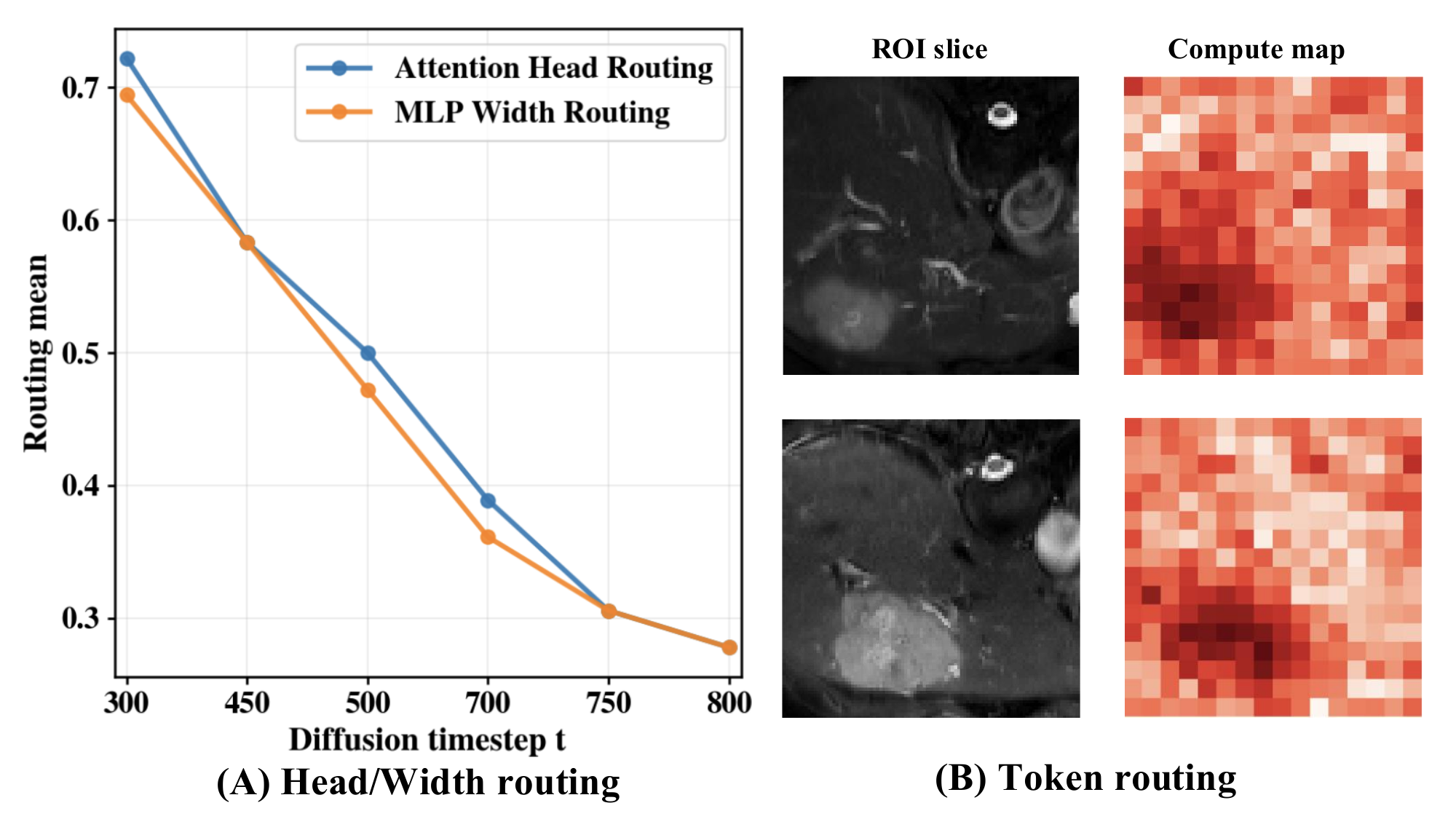}
\caption{Routing analysis. (A) Average head and width routing ratios across timesteps. (B) Token-wise computation map.
}
\label{fig:routing}
\end{figure}


As shown in Figure~\ref{fig:routing}, the attention head utilization decreases from 0.72 to 0.27, 
and the MLP width utilization from 0.69 to 0.27 as the timestep increases, indicating reduced computational demand at higher timesteps. 
Furthermore, token-wise computation is concentrated on specific regions, showing that the model selectively allocates computation to informative areas.

\subsection{Ablation Study}


\begin{table}[!t]
\centering
\caption{Ablation study on routing components.}
\label{tab:ablation}
\setlength{\tabcolsep}{6pt}
\begin{tabular}{lccc}
\hline
Ablation & FLOPs (G) & Lat.\ (ms) & AUC \\
\hline
\multicolumn{4}{l}{\textit{--- Module Contribution ---}} \\
w/o Head routing       & 338.28 & 222.62 & 0.725 \\
w/o Width routing      & 294.58 & 176.23 & 0.719 \\
w/o Token selection    & 273.30 & 155.26 & 0.698 \\[2pt]
\multicolumn{4}{l}{\textit{--- Token Selection Design ---}} \\
w/o Local conv         & 245.07 & 129.14  & 0.700 \\
\hline
\textbf{DiT-RC}   & \textbf{257.57} & \textbf{140.47} & \textbf{0.731} \\
\hline
\end{tabular}
\end{table}

\begin{table}[!t]
\centering
\caption{Effect of the budget loss coefficient $\lambda_{\mathrm{target}}$.}
\label{tab:target}
\footnotesize
\setlength{\tabcolsep}{4pt}
\begin{tabular}{c c c c}
\hline
$\lambda_{\mathrm{target}}$ & FLOPs (G) & Lat. (ms) & AUC \\
\hline
0.3 & 114.51 & 95.71 & 0.650 \\
\textbf{0.5(Ours)} & \textbf{257.57} & \textbf{140.47} & \textbf{0.731} \\
0.7 & 341.23 & 225.91 & 0.733 \\
1.0 & 418.80 & 321.23 & 0.740 \\
\hline
\end{tabular}
\end{table}
Table~\ref{tab:ablation} shows that head routing has the largest impact on performance, while token selection is essential for spatially adaptive computation. Removing local convolution slightly degrades performance, indicating the importance of local context. Overall, combining global and local routing yields the best performance–efficiency balance.


Table~\ref{tab:target} shows that the routing budget controls the trade-off between performance and efficiency.
Smaller budgets reduce computation but degrade performance, while larger budgets improve performance at higher cost.
Our setting achieves a balanced trade-off. Notably, $\lambda_{\text{target}}=1.0$ achieves higher AUC than DiT-C at the same FLOPs. As routing is effectively inactive in this setting, both models share the same inference structure. The improvement is attributed to the regularization effect of the budget loss.

\section{Conclusion}
This work presented an efficient diffusion transformer classifier for preoperative PNI prediction from volumetric MRI. 
By introducing adaptive routing across attention heads, spatial tokens, and MLP width, along with a local context-aware convolution module, the proposed method dynamically allocates computation according to input complexity. Experimental results demonstrate that the model achieves 0.731 AUC at 257.57 GFLOPs, 
while maintaining stable predictive performance.
These findings demonstrate the potential of diffusion transformer-based classification with adaptive computation for efficient and robust medical image analysis, although external multi-center validation remains necessary to assess generalization across institutions, scanners, and imaging protocols.

\section*{Acknowledgment}
This work was supported by the IITP grant 
(IITP-2023-RS-2023-00256081) funded by MSIT, Korea, and the ANCHOR program 
(2026-ANCHOR-01-110) funded by the Ministry of Education and the Seoul 
Metropolitan Government, Republic of Korea.

\bibliographystyle{ieeetr}
\bibliography{myref}

\end{document}